\documentclass[a4paper,fleqn]{cas-sc}

\usepackage[authoryear]{natbib}
\usepackage{subcaption}
\def\tsc#1{\csdef{#1}{\textsc{\lowercase{#1}}\xspace}}
\tsc{WGM}
\tsc{QE}

\newcommand{\model}{\textsc{MineROI-Net}}

\begin{document}
\let\WriteBookmarks\relax
\def\floatpagepagefraction{1}
\def\textpagefraction{.001}

\shorttitle{Smart Timing for Mining}    

\shortauthors{S. Wickramasinghe et al.}  

\title [mode = title]{Smart Timing for Mining: A Deep Learning Framework for Bitcoin Hardware ROI prediction}  



%
\author[1]{Sithumi Wickramasinghe}[type=author,
      style=english,
      auid=000,
      bioid=1,
      prefix=,
      orcid=0009-0004-6641-852X,
      ]


\cormark[1]


\ead{sithumi_wickramasinghe@mymail.sutd.edu.sg}


\credit{Conceptualization, Methodology, Software, Validation, Formal analysis, Visualization, Writing - Original Draft}

\affiliation[a]{organization={Singapore University of Technology and Design},
            addressline={8 Somapah Road}, 
            city={Singapore},
            postcode={487372}, 
            country={Singapore}}

\author[1]{Bikramjit Das}[type=author,
      style=english,
      auid=000,
      bioid=1,
      prefix=,
      orcid=0000-0002-6172-8228
      ]
\ead{bikram@sutd.edu.sg}    
\credit{Supervision, Writing - Review \& Editing}

\author[1]{Dorien Herremans}[type=author,
      style=english,
      auid=000,
      bioid=1,
      prefix=,
      orcid=0000-0001-8607-1640
      ]


\ead{dorien_herremans@sutd.edu.sg}

\credit{Supervision, Writing - Review \& Editing}


\cortext[1]{Corresponding author}



\begin{abstract}
Bitcoin mining hardware acquisition requires strategic timing due to volatile markets, rapid technological obsolescence, and protocol-driven revenue cycles. Despite mining's evolution into a capital-intensive industry, there is little guidance on when to purchase new Application-Specific Integrated Circuit (ASIC) hardware, and no prior computational frameworks address this decision problem. We address this gap by formulating hardware acquisition as a time series classification task, predicting whether purchasing ASIC machines yields profitable (Return on Investment $\text{(ROI)} \geq 1$), marginal ($0 < \text{ROI} < 1$), or unprofitable ($\text{ROI} \leq 0$) returns within one year. We propose \model{}, an open-source Transformer-based architecture designed to capture multi-scale temporal patterns in mining profitability. Evaluated on data from 20 ASIC miners released between 2015 and 2024 across diverse market regimes, \model{} outperforms recurrent, convolutional, and attention-based baselines, achieving 83.2\% accuracy and 83.5\% macro F1-score. The model demonstrates strong economic relevance, achieving 97.8\% precision in detecting unprofitable periods and 81.5\% precision in detecting profitable ones, while avoiding misclassifying profitable scenarios as unprofitable and vice versa. These results indicate that \model{} offers a practical, data-driven tool for timing mining hardware acquisitions, potentially reducing financial risk in capital-intensive mining operations.
\end{abstract}




\maketitle

\section{
Introduction
}
\label{introduction}
Bitcoin, introduced by \cite{nakamoto2008bitcoin}, established a peer-to-peer electronic cash system secured through a proof-of-work consensus. By enabling direct digital transactions without intermediaries, it laid the foundation for decentralized financial systems and sparked the growth of the broader cryptocurrency economy \citep{bohme2015bitcoin, catalini2020some}. Beyond its role as a payment network, Bitcoin represents a socio-technical experiment in digital scarcity: a fixed supply of 21 million coins \citep{hayes2017cryptocurrency, pagnotta2018equilibrium}, issued according to a predetermined algorithmic schedule, ensures that its value is shaped by transparent protocol rules rather than centralized control.

At the heart of Bitcoin's security lies mining, the computational process through which new coins are issued while validating transactions and securing the distributed ledger \citep{nakamoto2008bitcoin, courtois2013unreasonable}. Miners compete to solve cryptographic puzzles by repeatedly evaluating the SHA-256 proof-of-work function \citep{bonneau2015sok}, thereby converting electricity into probabilistic chances of earning block rewards and transaction fees. Bitcoin's issuance schedule includes periodic halvings that reduce the block reward by approximately 50\% every 4 years \citep{nakamoto2008bitcoin}. These halvings serve as a supply-control mechanism, but they also introduce cyclical shocks to miner revenue, as empirical studies show that miners often anticipate halvings by upgrading equipment or scaling operations ahead of reward reductions \citep{jablczynska2023energy}. Over time, mining has evolved from CPUs and GPUs to field-programmable gate arrays (FPGAs) and, ultimately, highly specialized application-specific integrated circuits (ASICs) designed exclusively for hashing computations \citep{courtois2013unreasonable}. These ASICs deliver several orders of magnitude of improvement in energy efficiency and computational throughput, but they also depreciate rapidly as newer models enter the market, making mining a capital-intensive global industry where profitability depends on Bitcoin's market price, network difficulty, hardware efficiency, and electricity costs \citep{hayes2017cryptocurrency, jablczynska2023energy}.

Despite mining's critical role in Bitcoin's operation, the strategic timing of hardware acquisitions remains poorly understood and underexplored in the literature. The combination of hardware obsolescence, electricity price volatility, and protocol-driven revenue cycles makes purchase timing one of the most critical decisions for mining operators; yet, no formal framework exists to guide when to expand capacity. Most prior research has focused on optimizing existing mining operations by applying machine learning to improve efficiency and reduce energy consumption or examining miners' optimal exit and hardware sale timing under changing market conditions \citep{li2024sale,jablczynska2023energy}. However, the complementary question of when to acquire new hardware to maintain a competitive advantage has received little attention. This gap is particularly problematic, given that mining profitability exhibits strong cyclicality around halving epochs and market cycles, which complicates investment decisions.

The consequences of poorly timed hardware investments can be severe and persistent. Industry analyses illustrate how purchasing decisions made during market peaks can significantly damage returns: miners who purchased Antminer S19j Pro hardware during the 2021 bull-market peak saw their expected payback period increase from approximately 13 months to 107 months after mining margins collapsed \citep{hashrateindex2022timing}. Such miscalculations can render entire operations economically unviable, as operators must continue to bear electricity costs while hardware depreciates and newer, more efficient models enter the market. In the absence of data-driven decision frameworks, miners often rely on heuristics or intuition, often entering the market precisely when conditions are least favorable, a classic example of pro-cyclical investment behavior that amplifies financial risk. The lack of systematic approaches to hardware acquisition timing, therefore, represents a fundamental gap in mining economics research and industry practice.

To address this gap, we formulate mining hardware acquisition as a multi-class time-series classification problem. We construct a comprehensive dataset spanning 20 ASIC mining machine types from 2015 to 2024 by integrating machine-specific attributes, Bitcoin market and price indicators, and Bitcoin network-level variables. We train and evaluate models that are based on the Long-short-Term Memory (LSTM) \citep{hochreiter1997long}, Transformer \citep{vaswani2017attention}, and TSLANet \citep{eldele2024tslanet} architectures to classify whether a hardware purchase on a given day yields a profitable Return on Investment ($ROI \geq 1$), a moderate return ($0 < ROI < 1$), or a negative return ($ROI \leq 0$) within a 365-day horizon. This multi-class, data-driven framework provides actionable decision support for mining operators by predicting profitability outcomes.

Our main contributions are as follows: (i) We introduce the novel task of Bitcoin mining hardware acquisition timing, framing it as a multi-class time series classification problem; (ii) to enable this new task, we propose a method for collating a unique dataset from various sources, making it aptly suited for studying mining economics; and finally, (iii) we propose the \model{} model that achieves 83.2\% accuracy with 97.8\%/81.5\% precision for unprofitable/profitable predictions and was benchmarked against an LSTM-based, transformer-based baseline and TSLANet architectures. The remainder of this paper is organized as follows: first, in Section~\ref{sec:related} we review related work, then, in Section~\ref{sec:Methodology} we detail our approach and model, followed by our experimental setup and results in Section~\ref{sec:Experimental_setup} and Section~\ref{sec:Results}.  Finally, Section~\ref{sec:Conclusion} concludes the paper.

\section{
Related work
}
\label{sec:related}

\subsection{
Mining economics and profitability drivers
}\label{subsec:Mining_economics}

A substantial body of research models bitcoin mining profitability as a function of bitcoin price, network difficulty, hardware efficiency, and electricity costs. \cite{delgado2019bitcoin} shows that profitability constraints can generate sharp economic thresholds. Using updated estimates of network energy consumption and production costs, the authors conclude that, by mid-2018, mining had become unprofitable for commodity miners without access to electricity priced below \$0.14 kWh, thereby encouraging geographic concentration in low-cost regions such as China, Iran, Russia, and Ethiopia. Ethiopia has more recently attracted attention for its mining industry with hydroelectricity rates as low as 3.2 cents per kWh~\citep{coinreporter2025ethiopia}. Complementary empirical and modeling studies suggest that mining profitability tends toward a competitive equilibrium, in which margins compress as participants enter and exit in response to rewards and costs. \cite{derks2018chaining} analyzes the actors and the value flows between these actors involved in the bitcoin system from January 2012 to December 2016 and reports that the marginal profit of mining can become negative, caused by the consensus mechanism of the bitcoin protocol, which requires a substantial investment in hardware and significant recurring daily expenses for energy.  

Several studies emphasize the primacy of energy costs as the key driver of operating expenses, while also noting that overhead (e.g., cooling/infrastructure) and hardware procurement/renewal matter but are harder to observe consistently \citep{song2020cost, sapra2025powering}. \cite{song2020cost} explicitly decomposes mining cost into (1) energy cost, (2) facility overheads, and (3) purchasing/renewing hardware, but they focus on the energy cost of running the bitcoin mining hardware as likely to be the key driver and the only component estimable with reasonable precision in public data. Recent evidence continues to reinforce this view. The Cambridge Digital Mining Industry Report shows that electricity accounts for more than 80\% of miners' cash-based operating expenses, while also documenting continued improvements in deployed ASIC hardware efficiency \citep{cambridge2025mining}. In turn, \cite{hayes2017cryptocurrency}' cost-of-production model links Bitcoin’s implied production value to miners’ marginal costs-primarily electricity price, hardware energy efficiency, and network difficulty-providing a break-even/lower-bound perspective that motivates treating electricity price as a key variable in mining decision support. \cite{sapra2025powering} also supports treating the electricity price as a key variable in mining decision support. Beyond cost structure, miners’ revenue is strongly shaped by protocol-level mechanics (block rewards and halvings) and fee-market dynamics. 

Beyond descriptive cost analyses, a growing body of work develops formal economic models of the mining industry. \cite{haliplii2020mine} apply real option theory to simulates the fundamental mining reward and measures the likelihood of breakeven on initial investment, revealing that miners failed to adjust to price signals after the 2017 bitcoin bubble. In parallel, research has examined the incentive mechanisms governing transaction fees and miner-user interactions. \cite{hiraide2023analysis} develops a queueing and game-theoretic (Nash equilibrium) model linking transaction fees, confirmation delays, newly issued coins, and total hash rate. Recent empirical work further suggests that hashrate is not merely a technical network indicator but also an economically meaningful variable, as it both responds to mining revenue incentives and significantly affects miners’ revenues and mining-related resource use \citep{hu2025time}. Such analyses underscore the importance of fee dynamics and network-level indicators in determining mining revenue and motivate their inclusion in predictive profitability models. More recently, \cite{prayoga2025bitcoin} employed boosting ensemble methods to predict daily mining device income using 60-day windows of market and machine-level features across 70 ASIC machines released between 2020 and 2024 (4,200 samples). While their approach demonstrates the feasibility of machine learning for mining-income prediction, the study focuses on continuous revenue estimation rather than the capital investment decision itself. In particular, it does not incorporate hardware acquisition costs, does not define an explicit ROI horizon, such as a 1-year cost recovery period, and does not evaluate model generalization under temporal regime shifts or the introduction of new hardware. Consequently, the problem of acquisition timing under capital commitment remains largely unexplored.

\subsection{
Deep learning for cryptocurrency markets and time-series decision support
} \label{subsec:Deep_learning}

Cryptocurrency markets exhibit strong nonlinearity, regime shifts, and multi-scale temporal dependencies, which make traditional linear econometric models such as ARIMA insufficient for capturing complex dynamics \citep{quang2025analysis,kamal2025predicting,yeganeh2026monitoring}. As a result, deep learning architectures have become the dominant paradigm for cryptocurrency price forecasting, volatility modeling, and trend prediction \citep{kamal2025predicting,han2025mfb,oyedele2023performance,guresen2011using}.

Recurrent neural networks, particularly Long Short-Term Memory (LSTM) \citep{hochreiter1997long} models, remain important baseline architectures due to their ability to model sequential dependencies. Survey evidence indicates that LSTM-based models consistently outperform classical statistical approaches in cryptocurrency forecasting tasks~\citep{zhang2024survey}. Empirical studies further refine this direction: ~\cite{seabe2023forecasting} demonstrate that Bi-LSTM variants reduce prediction error across multiple cryptocurrencies, while ~\cite{chen2025cryptocurrency} propose multi-task LSTM architectures that jointly address price forecasting and portfolio optimization. Hybrid extensions such as MRC-LSTM~\citep{guo2021mrc} integrate multiscale residual convolutional blocks with recurrent units to capture both short-term fluctuations and longer-term structural trends. Beyond pure price-based modeling, multimodal approaches incorporate external information sources. The PreBit framework \citep{zou2023prebit} combines FinBERT-based sentiment embeddings from Twitter with technical indicators through a CNN–SVM architecture to predict extreme Bitcoin price movements. Recent work further strengthens this direction by showing that integrating financial variables with blockchain, news, and social-media information can improve cryptocurrency forecasting accuracy as well as portfolio performance \citep{gurgul2025deep,tabe2026proposing,han2025mfb}. These studies highlight the growing recognition that cryptocurrency markets are shaped by heterogeneous signals spanning technical, textual, and behavioral domains.

Transformer architectures further advance this line of research by leveraging self-attention mechanisms that model long-range dependencies and multivariate interactions \citep{vaswani2017attention}. Time-series adaptations of Transformers have demonstrated promising performance in volatile financial settings. For instance, the Synthesizer Transformer in \cite{herremans2025forecasting} effectively captures volatility spikes and outperforms several traditional econometric baselines, and \cite{fischer2024fx} shows that Transformer architectures with temporal embeddings outperform traditional models in financial time-series prediction by better encoding time-dependent patterns. Similarly, multi-resolution, multi-scale, and patch-based Transformer variants have been proposed to explicitly capture hierarchical temporal dynamics, which are critical in non-stationary environments \citep{shao2025energyformer,zhang2025integrative,jin2024novel,nie2023patchtst,abeywickrama2025entrope}. Additionally, convolutional architectures have gained attention as computationally efficient models capable of effectively capturing local temporal patterns in time series data. TSLANet (Time Series Lightweight Adaptive Network) \citep{eldele2024tslanet} is a notable work that exemplifies this approach, combining Adaptive spectral blocks with interactive convolution blocks to capture both long-term and short-term patterns. It balances representational power and computational efficiency, achieving competitive results across time-series forecasting and classification benchmarks. More broadly, recent review evidence suggests that the literature is increasingly moving from standalone recurrent models toward Transformer-based, hybrid, and multimodal frameworks for cryptocurrency forecasting \citep{wu2026review}. 

Despite these advances, prior deep learning research in the cryptocurrency domain has focused primarily on price forecasting, return prediction, volatility modeling, and directional classification. Existing studies do not formulate mining hardware acquisition as a classification problem aligned with capital commitment and cost recovery over a predefined ROI horizon. In contrast, our work formulates mining hardware timing as a time-series classification task, in which the target variable explicitly integrates hardware purchase price, electricity cost assumptions, network dynamics, and a fixed one-year investment horizon. By transforming profitability analysis into a decision-support problem, our approach bridges deep learning-based temporal modeling with capital allocation strategy in capital-intensive cryptocurrency mining operations.

\section{
Approach and model
} \label{sec:Methodology}
We formulate the Bitcoin mining hardware acquisition timing problem as a multi-class time-series classification task. Given a sequence of historical observations up to day \(d_i\), the objective is to predict the one-year Return on Investment (ROI) category for purchasing a specific mining machine $M$ on that day. Let
\(\mathcal{D}=\{(X^M_i,y^M_i)\}_{i=1}^{N}\) denote our dataset, where \(X^M_i\in\mathbb{R}^{L\times F}\) represents a multivariate time series window of length \(L\) (look-back period) with \(F\) features, \(y^M_i\in\{0,1,2\}\) is the discrete ROI label for purchasing machine \(M\) on day \(d_i\) and \(N\) is the total number of samples (time windows). Each sequence \(X^M_i\) contains the past $L$-day window of machine-specific attributes, market indicators, and network-level features up to day \(d_i\). These features are discussed in more detail in Section~\ref{sec:dataset}.

\subsection{
ROI computation and class labeling
}
\label{subsec:roi_labeling}
We model the acquisition of an ASIC machine $M$ on purchase date $d_i$ as a one-year cash flow process with a horizon of $H=365$ days. Let $t \in \{1, \ldots, H\}$ denote the number of days elapsed since $d_i$, so that all time-varying quantities indexed by $(i,t)$ refer to the value observed $t$~days after the purchase corresponding to sample~$i$.
Let $C_M(d_i)$ denote the acquisition cost of machine $M$ in USD, $h_M$ the machine hash rate (H/s), and $P_M$ the electrical power consumption (W) of machine $M$. 
Following common operational practice among miners, we assume that mined Bitcoin is periodically converted to fiat currency to cover operational costs; therefore, daily mining revenue is evaluated in USD using the prevailing BTC/USD\footnote{The US dollar will be used to price bitcoin solely for illustrative purposes. In actuality, there are bitcoin miners all over the world, who will purchase electricity at their local rate and in their local currency.} market price.

\paragraph{Daily mined BTC:}
At elapsed day~$t$ after purchase date~$d_i$, the expected BTC mined by machine $M$ is modeled as
\begin{equation}
m_{i,t}(M) = 
\frac{h_M \cdot 86400}{D_{i,t} \cdot 2^{32}}
\left(B_{i,t} + \frac{F_{i,t}}{144}\right),
\label{eq:mined_btc}
\end{equation}
where $D_{i,t}$ denotes network mining difficulty, $B_{i,t}$ the block subsidy (BTC/block), and $F_{i,t}$ the total transaction fees accrued per day. Since approximately 144 blocks are produced daily, fees are normalized per block as $F_{i,t}/144$. The constant $2^{32}$ relates to the normalized probability of a single hash per second solving a block, and is a feature of the 256-bit encryption at the core of the SHA-256 hashing function \citep{hayes2015decision}. Daily gross revenue in USD is therefore, $p_{i,t} \cdot m_{i,t}(M)$, where $p_{i,t}$ denotes the market price of Bitcoin in USD on day $t$.

\paragraph{Exchange fee:}
Mining revenue is typically realized through cryptocurrency exchanges, where mined Bitcoin is converted into fiat currency. These exchanges charge transaction fees that depend on trading volume, account tier, and order type. Across major platforms such as Binance, Kraken, and OKX, base trading fees are commonly close to 0.10\% per transaction for standard users \footnote{\url{https://www.binance.com/en/fee/trading}}\footnote{ \url{https://www.okcoin.com/fees.html}}\footnote{\url{https://www.kraken.com/features/fee-schedule}}. Because most small and mid-scale mining operators do not qualify for institutional pricing tiers, we adopt a representative trading fee factor $\phi$ = $0.001$. Accordingly, the net daily mining revenue is computed as
\begin{equation}
R^{\text{net}}_{i,t}(M) = (1-\phi) \cdot p_{i,t} \cdot m_{i,t}(M)
\label{eq:net_revenue}
\end{equation}

\paragraph{Electricity cost:}
\label{subsubsection:Electricity_cost}
The daily electricity expenditure for a machine $M$ can be modeled as
\begin{equation}
C^{\text{elec}}_{i,t}(M) = \frac{P_M \cdot 24}{1000}\, r_e,
\label{eq:elec_cost}
\end{equation}
where $P_M$ the electrical power consumption (W) and $r_e$ is the electricity tariff (USD/kWh). We evaluate mining profitability under a set of constant electricity-rate scenarios (0.01-0.20 USD/kWh). This design reflects that industrial consumers can lock in electricity procurement under contractual arrangements such as power purchase agreements (PPAs), which enable purchasing energy at a fixed price for a specific duration \citep{nour2022blockchain_electricity_review}. We therefore treat the electricity rate as fixed within each scenario during ROI computation.

\paragraph{Maintenance cost:}
ASIC miners require periodic maintenance, including cleaning, cooling system servicing, and replacement of degraded components. Industry reports and operational analyses suggest annual maintenance expenditure typically ranges between 5-10\% of initial equipment cost 
, depending on operating scale and environmental conditions. Smaller operations often fall near the lower bound of this range. To obtain a conservative and reproducible estimate, we assume a maintenance costs equal to 5\% ($\alpha=0.05$) of the initial machine purchase price per year, amortized uniformly over the horizon:
\begin{equation}
C^{\text{maint}}_{i,t}(M) = \frac{\alpha \cdot C_M(d_i)}{365}.
\label{eq:maint_cost}
\end{equation}

\paragraph{Daily profit and cumulative ROI:}
Having defined the revenue and cost components of mining operations, we now formalize the profit and return-on-investment measures used in this study. All revenues and costs are expressed in U.S. dollars (USD). For a machine $M$ purchased on date $d_{i}$, the daily net profit at time $t$ is defined as net mining revenue minus electricity and maintenance costs:
\begin{equation}
\pi_{i,t}(M) = R^{\text{net}}_{i,t}(M) - C^{\text{elec}}_{i,t}(M) - C^{\text{maint}}_{i,t}(M).
\label{eq:daily_profit}
\end{equation}
The cumulative profit over the one-year horizon $H$ is then computed by aggregating daily profit across all days in the evaluation period:
\begin{equation}
\Pi_i(M) = \sum_{t=1}^{H} \pi_{i,t}(M),
\end{equation}
Based on this cumulative profit, the one-year return on investment is defined as
\begin{equation}
\mathrm{ROI}(M,d_i) = \frac{\Pi_i(M)}{C_M(d_i)}.
\label{eq:roi}
\end{equation}
This formulation expresses profitability relative to the initial hardware purchase cost, thereby providing a direct measure of capital recovery over the investment horizon. The baseline analysis assumes daily conversion of mined Bitcoin into fiat currency; however, because conversion timing can materially affect realized profitability under volatile market conditions, alternative liquidation schedules, including monthly conversion, are examined separately in the Section \ref{subsec:selling_strategies}.

\paragraph{ROI-based class labeling:}
To enable supervised learning, continuous ROI values are discretized into economically interpretable categories:
\begin{equation}
y_i=
\begin{cases}
0, & \mathrm{ROI}(M,d_i)\le 0 \quad \text{(unprofitable)},\\
1, & 0<\mathrm{ROI}(M,d_i)<1 \quad \text{(marginal)},\\
2, & \mathrm{ROI}(M,d_i)\ge 1 \quad \text{(profitable)}.
\end{cases}
\label{eq:labeling}
\end{equation}
These correspond to, Class 0: capital loss, Class 1: partial capital recovery, Class 2: full capital recovery with at least 100\% return within one year. This categorization yields decision-oriented labels aligned with investment planning and hardware acquisition strategies. In future research, we may explore different ways to bin ROI, or even directly predict ROI with a regression approach.

\subsection{
Proposed model architecture
}
\label{subsec:model_architecture}
\begin{figure*}
    \centering
    \includegraphics[width=\linewidth]{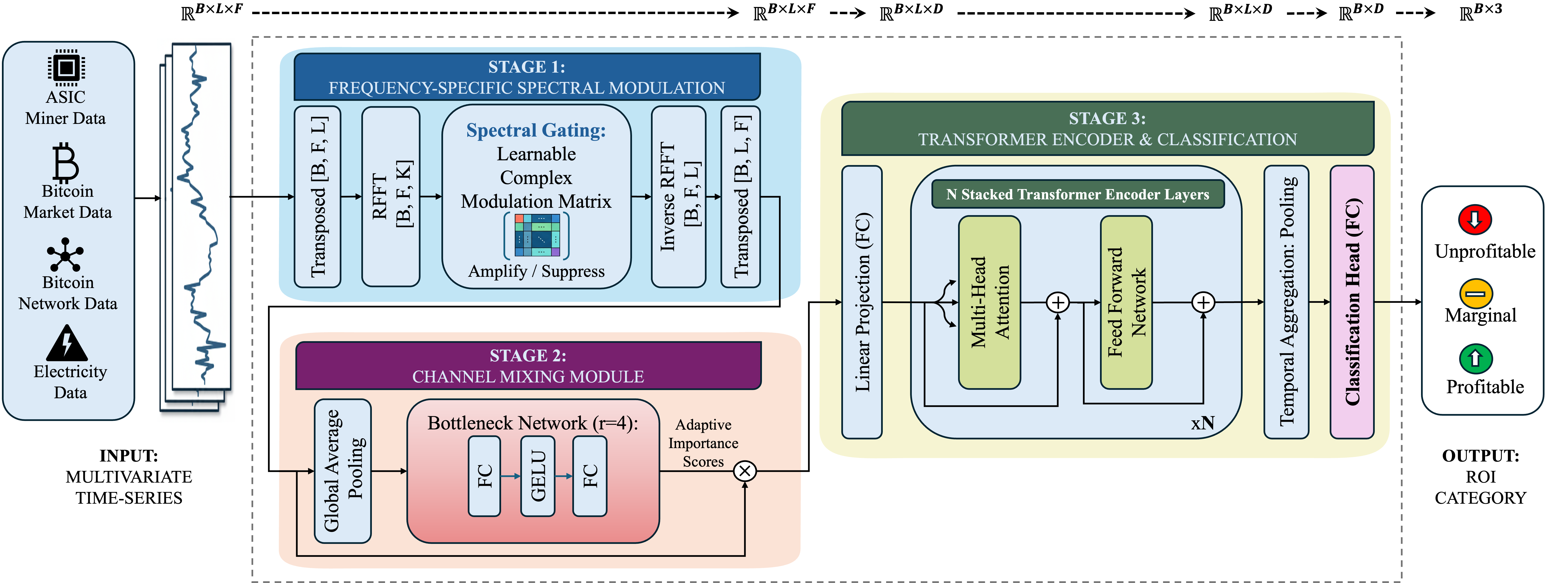}
    \caption{\model{} architecture overview.}
    \label{fig:Transformer_model}
\end{figure*}

To predict the ROI of a Bitcoin Mining machine purchased at a given time, it is essential to account for temporal dependencies, regime shifts, and multiscale patterns spanning daily fluctuations to multi-year cycles. Hence, deep learning models are better suited compared to traditional methods. We propose a unified deep learning architecture for predicting ROI for mining hardware, which we term \model{}. Given an input sequence $\mathbf{X} \in \mathbb{R}^{B \times L \times F}$, where $B$ is batch size, $L \in \{30, 60\}$ is the look-back window in days, and $F$ is the number of features, this model processes sequences through three main components as shown in Figure \ref{fig:Transformer_model}: the Spectral Feature Extractor, the Channel Mixing module and the Transformer Encoder.

\subsubsection{
Frequency-Specific Spectral Modulation
}
\label{subsubsec:Spectral_Modulation}
Bitcoin mining profitability is driven by cyclical dynamics that operate across multiple timescales: halving events recur approximately every four years;  difficulty adjustments occur every two weeks; and market sentiment cycles can span months. These periodic structures are not easily captured by standard transformer encoders, which process input sequences point-by-point in the time domain and rely on self-attention to implicitly discover temporal dependencies. While self-attention can capture long-range dependencies, it has no explicit mechanism for periodic structure and must learn cyclical patterns implicitly from data, making it data-hungry and potentially brittle under distribution shifts \citep{zeng2023transformers, nie2023patchtst}.

To address this limitation, motivated by frequency-enhanced time-series models \citep{fedformer,yi2023freq}, we incorporate a lightweight spectral block that transforms input sequences into the frequency domain using the Fast Fourier Transform (FFT), applies learnable frequency domain modulation, and maps the result back to the time domain via the inverse FFT. By operating explicitly in the frequency domain, the module enables the model to selectively amplify or suppress specific periodic components on a per-feature basis, providing an inductive bias toward cyclical structure that complements the transformer's attention-based temporal modeling.

Given an input sequence \(X \in \mathbb{R}^{B \times L \times F}\), where \(B\) is batch size, \(L\) is sequence length (look-back window), and \(F\) is the number of features, we first transpose \(X\) to \(X^\top \in \mathbb{R}^{B \times F \times L}\) so that each feature channel can be transformed independently along the temporal axis using a real FFT (RFFT) :

\begin{equation}
\hat{X} = \mathrm{RFFT}\!\left(X^\top\right), \qquad \hat{X} \in \mathbb{C}^{B \times F \times K},
\end{equation}
where \(K=\lfloor L/2 \rfloor + 1\) denotes the number of non-redundant frequency bins for real-valued signals. We then learn a frequency-specific complex modulation matrix \(W \in \mathbb{C}^{F \times K}\) and perform element-wise spectral gating:

\begin{equation}
\tilde{X}_{b,f,k} = W_{f,k} \cdot \hat{X}_{b,f,k}, \quad
\forall b \in [1,B],\; f \in [1,F],\; k \in [1,K].
\label{eq:spectral_modulation}
\end{equation}

Finally, the modulated spectrum is mapped back to the time domain via the inverse real FFT:
\begin{equation}
{\mathrm{X}_{spec}} = \mathrm{IRFFT}\!\left(\tilde{X}\right)^\top, \qquad
{\mathrm{X}_{spec}} \in \mathbb{R}^{B \times L \times F}.
\end{equation}

This design enables channel-wise \emph{and} frequency-bin-wise modulation, allowing the model to amplify or attenuate informative periodic components while suppressing less useful spectral content.

\subsubsection{
Channel Mixing Module
} 
\label{subsubsec:Channel_Mixing}

While the Spectral Feature Extractor operates within each feature independently, identifying which temporal frequencies are informative and it does not model relationships across features. In practice, the relative importance of features varies substantially across market regimes. A model that treats all features equally regardless of context will fail to adapt to these regime-dependent dynamics.

To capture cross-feature interactions that vary across regimes (e.g., price vs. electricity cost dominance), we employ a channel mixing module inspired by Squeeze-and-Excitation networks~\citep{hu2018squeeze}. Unlike cross-attention, which computes pairwise feature interactions, this mechanism compresses temporal information via global average pooling, learns channel-wise attention weights via a bottleneck network, and reweights features accordingly. Recent work has demonstrated that efficient channel-mixing architectures can achieve competitive performance compared to attention mechanisms in time series forecasting~\citep{chen2023tsmixer,nie2023patchtst}.

We first collapse the temporal dimension via global average pooling to obtain a single context vector summarizing the entire input window for each feature. This pooling operation discards fine-grained temporal detail and retains only the global statistical summary of each feature over the window, providing a compact representation of the current market regime.
\begin{equation}
\mathbf{z} = \frac{1}{L}\sum_{t=1}^{L} \mathbf{X}_{\text{spectral}}[:, t, :] \in \mathbb{R}^{B \times F} 
\end{equation}
The pooled vector $z$ is passed through a two-layer fully connected bottleneck network with a reduction ratio $r=4$:
\begin{equation}
\mathbf{s} = \text{FC}_2(\text{GELU}(\text{FC}_1(\mathbf{z}))) \in \mathbb{R}^{B \times F}
\end{equation}
where $\text{FC}_1 : \mathbb{R}^{F} \rightarrow \mathbb{R}^{F/r}$ compresses the feature vector from $F$ to $F/r$ dimensions, and $\text{FC}_2 : \mathbb{R}^{F/r} \rightarrow \mathbb{R}^{F}$ projects back to $F$ dimensions. Each $\text{FC}(\cdot)$ denotes a fully connected layer comprising a learnable weight matrix and bias vector. The GELU non-linearity enables the network to model non-linear interactions between features. Critically, the bottleneck forces all $F$ features to interact through a shared compressed representation before importance scores are computed, ensuring that each feature's importance score is influenced by the global context of all other features rather than being computed in isolation. The output vector $s$ serves as a set of $F$ channel-wise importance scores, one per feature. These scores are broadcast across the time dimension and applied multiplicatively to the output of the spectral modulation stage:
\begin{equation}
    \mathbf{X}_{\text{mixed}} = \mathbf{X}_{\text{spectral}} \odot \mathbf{s} \in \mathbb{R}^{B \times L \times F}
\end{equation}
Features assigned high importance scores are amplified, while less relevant features are suppressed, producing a context-adaptive representation that emphasizes the signals most relevant to ROI prediction under the current market conditions.

\subsubsection{
Transformer Encoder
}
\label{subsubsec:Transformer_Encoder}
The spectrally enhanced and channel-reweighted representation $\mathbf{X}_{\text{mixed}}\in \mathbb{R}^{B \times L \times F}$ is next processed by a standard Transformer encoder \citep{vaswani2017attention}. Since the Transformer operates in a fixed latent dimension $D = d_{\text{model}}$, each $F$-dimensional input vector is first projected into this latent space $d_{\text{model}}$ using a position-wise fully connected layer, denoted by $FC(.)$. Sinusoidal positional encodings $\mathbf{PE}\in \mathbb{R}^{B \times L \times d_{\text{model}}}$  are then added to inject temporal order information. The resulting initial Transformer input representation, denoted by $Z_{0}$, is therefore given by
\begin{equation}
    \mathbf{Z}_0 = \mathrm{FC}(\mathbf{X}_{\text{mixed}}) + \mathrm{PE}
    \in \mathbb{R}^{B \times L \times d_{\text{model}}}
    \label{eq:transformer-input}
\end{equation}
The encoded representation $\mathbf{Z}_{0}$ is then passed through $N$ stacked Transformer encoder layers, each consisting of multi-head self-attention, a position-wise feed-forward network, layer normalization, and residual connections, to produce the final temporal representation used for classification.

\subsubsection{
Classification and Training
}
\label{subsubsec:Classification}
Let $\mathbf{Z}_{\text{N}}\in \mathbb{R}^{B \times L \times d_{\text{model}}}$ denote the output of the final Transformer encoder layer. To obtain a fixed-length representation for each sequence, we apply global average pooling over the temporal dimension. The pooled vector is then passed through a two-layer fully connected classification head, which outputs the predicted class probabilities $p_{i,c}$ for the three ROI classes. The model is trained using weighted cross-entropy loss with label smoothing, defined as
\begin{equation}
\qquad
\mathcal{L}(\theta)=-\frac{1}{N}\sum_{i=1}^{N}\sum_{c=1}^{3}w_c \cdot \tilde{y}_{i,c} \cdot \log p_{i,c}\
\end{equation}
\(\text{where} \
\tilde{y}_{i,c}=(1-\varepsilon)\,y_{i,c} + \varepsilon/3\). Here $N$ denotes the number of training samples, $i$ indexes individual samples, $c$ indexes 
the three ROI classes, $w_c$ are class weights computed via inverse frequency, 
$\tilde{y}_{i,c}$ is the smoothed target label with $\varepsilon = 0.1$, and $p_{i,c}$ 
is the predicted probability for sample $i$ and class $c$.

\section{
Experimental setup
}
\label{sec:Experimental_setup}
This section outlines the full experimental pipeline, covering the dataset construction, temporal data-splitting strategy, baseline implementations, and hyperparameter tuning methodology.

\subsection{
Dataset and input construction
}
\label{sec:dataset}
We construct a multivariate daily dataset that spans from October 2015 to September 2025. Since ROI labels require a one-year forward window to compute
(see Section~\ref{subsec:roi_labeling}), labeled samples are available only up to September 2024, covering multiple complete bitcoin market cycles, mining difficulty regimes, and ASIC hardware generations. Each sample corresponds to a candidate machine-purchase date $d_{i}$ and is represented by a look-back window of historical observations used to predict the one-year ROI class. The model's input combines machine-specific variables, Bitcoin market and network variables, and electricity-rate scenarios, corresponding to the four input groups illustrated in Figure \ref{fig:Transformer_model}.

\begin{enumerate}
    \item \textbf{ASIC Machine-data}: We collect hardware specifications for 20 machines from Hashrate Index\footnote{\url{https://data.hashrateindex.com/asic-index-data/price-index}}, a commercial mining data provider, and publicly available Amazon price tracker\footnote{\url{https://camelcamelcamel.com/}}. For each machine, we extract specification (hashrate, power requirements, efficiency, days since release date) and purchase price at various times. The machine set covers Antminer S-series machines (S7, S9, S15, S17 Pro, S19 variants, S21), T-series machines (T17, T19), WhatsMiner M-series machines (M10s-M53), and AvalonMiner KA3.
    
    \item \textbf{Blockchain Data}: We incorporate blockchain-related variables from Blockchain.com\footnote{\url{https://www.blockchain.com/}}. These variables include Bitcoin price, mining difficulty, network hashrate, network revenue, block reward, and transaction fees. We also include the number of days since the previous halving to expose the model to Bitcoin’s protocol-driven halving cycles and miners' daily revenue potential.

    \item \textbf{Energy prices}: To account for operating-cost variation, we evaluate profitability under fixed electricity-rate scenarios ranging from 0.01 to 0.20 USD/kWh, following the formulation in (\ref{eq:elec_cost}).
    
\end{enumerate}

To ensure temporal validity and prevent data leakage, all 14 features at day $d_i$ use only information available at or before $d_i$. All features are Min--Max normalized using statistics computed on the training split only, and the fitted scaler is applied unchanged to the validation and test splits.

\subsection{
Baseline models
}
\label{subsec:baselines}
We benchmark \model{} against a diverse set of deep learning baselines to evaluate the relative strengths of (i) sequential recurrence, (ii) attention-based long-range dependency modelling, and (iii) convolutional inductive biases for local pattern extraction. In addition, because mining profitability exhibits pronounced cyclicality (e.g., halving cycles, difficulty dynamics), we include frequency-aware designs to assess whether explicit spectral processing improves ROI classification beyond purely time-domain learning. Concretely, our baselines are:

\paragraph{Vanilla LSTM.} A standard stacked LSTM classifier that operates purely in the time domain, serving as a canonical sequential modelling baseline.

\paragraph{LSTM with blocks (LSTM + Spectral + Channel Mixing).} To disentangle the effect of the backbone from the input representation, we augment the LSTM with the same FFT-based Spectral Feature Extractor and Channel Mixing module used in \model{}, and then replace the Transformer encoder with stacked LSTM layers and a fully connected classification head. This controlled design allows us to test whether performance gains arise primarily from attention versus recurrence when the frequency-aware and cross-feature reweighting components are held constant.

\paragraph{Vanilla Transformer.} A standard Transformer encoder classifier operating directly in the time domain (with positional encoding and pooling-based aggregation), included to evaluate the benefit of attention-based modeling without any explicit spectral or channel-mixing modules.

\paragraph{TSLANet.} We include TSLANet (Time Series Lightweight Adaptive Network) \citep{eldele2024tslanet} as a strong CNN-based state-of-the-art baseline for time-series representation learning. TSLANet replaces self-attention with an Adaptive Spectral Block (ASB) that performs Fourier-domain processing (FFT/IFFT) with adaptive thresholding to suppress high-frequency noise, and combines it with an Interactive Convolution Block (ICB) that refines temporal representations using interacting convolutions with different kernel sizes. This baseline provides a principled comparison with a frequency-aware architecture that incorporates an explicit convolutional inductive bias for local dependency learning, while still capturing longer-range interactions through spectral filtering.

\paragraph{\model{} (Transformer + Spectral + Channel Mixing).} Our proposed model integrates an FFT-based spectral feature extractor with learnable complex weights and a channel-mixing module to emphasize ROI-relevant feature interactions, followed by a Transformer encoder to capture long-range temporal dependencies (Section \ref{subsec:model_architecture}).

\paragraph{Training protocol.} For fair comparison, all models are trained under the same optimization setup: batch size 64, up to 20 epochs, and AdamW \citep{loshchilov2017decoupled} optimizer with weight decay $1\times10^{-5}$; hyperparameters are selected using the expanding-window cross-validation procedure described in Section \ref{subsec:data_split}.

\subsection{
Time-based data split and expanding-window cross-validation
}
\label{subsec:data_split}
To preserve temporal ordering and prevent information leakage, we adopt a strictly time-based evaluation framework rather than random sampling throughout all stages of model development, including hyperparameter selection, model comparison, and final testing. 

\begin{figure*}
        \centering
        \includegraphics[width=\linewidth]{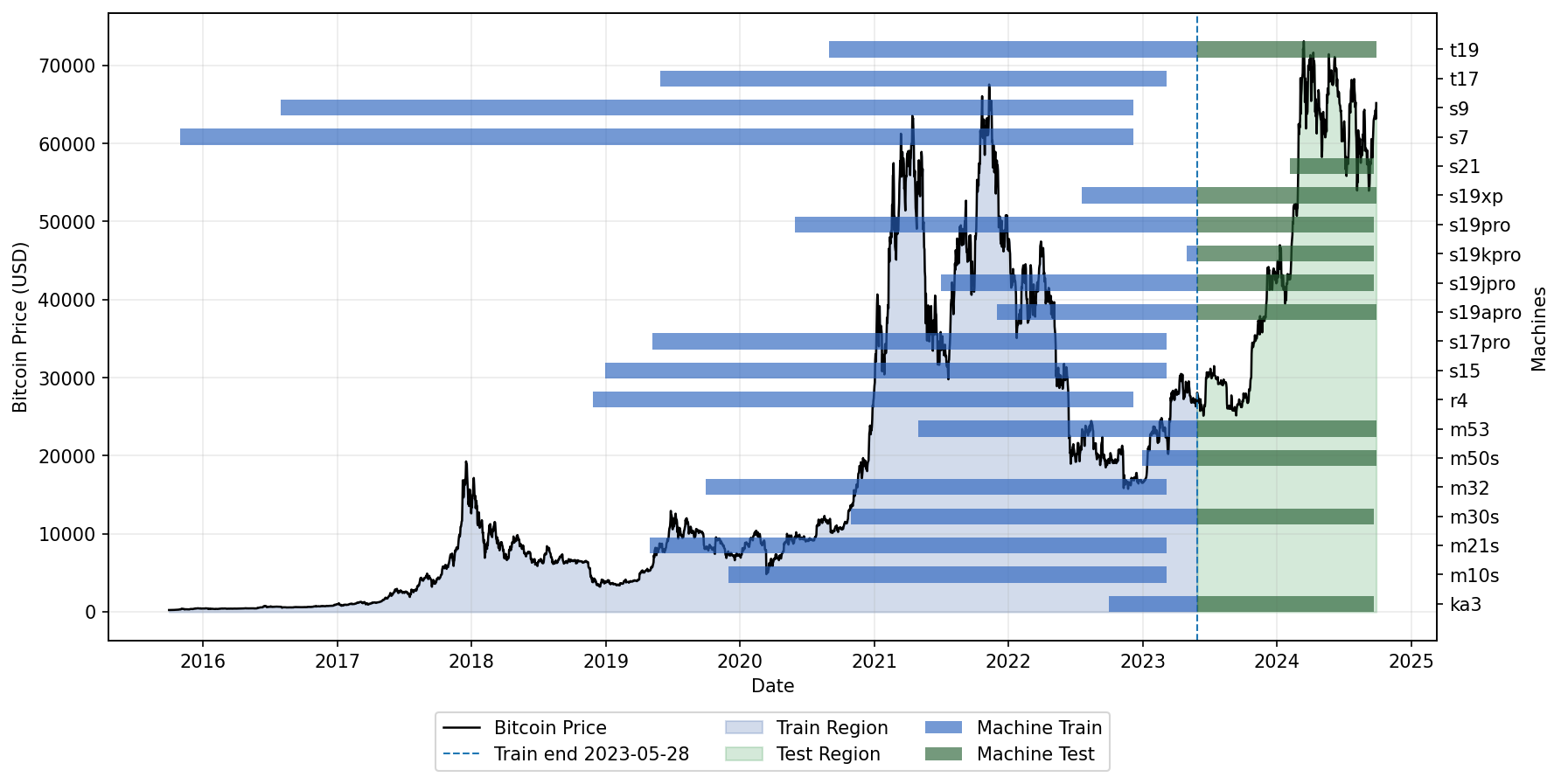}
        \caption{BTC price with train/test split and machine coverage}
        \label{fig:machine_splits}
\end{figure*}

\paragraph{Train-test partition:} The dataset is partitioned into training (80\%) and test (20\%) subsets based on the quantiles of the final window date. As illustrated in Figure~\ref{fig:machine_splits}, the training period spans from the beginning of the dataset up to May 2023, while the remaining observations are reserved for testing. Figure~\ref{fig:machine_splits} overlays the temporal distribution of training and test samples on the historical bitcoin price trajectory. In addition, the horizontal bars (right axis) depict the active periods of individual mining machines. Each machine contributes samples only during its availability window, and its sequences are consistently segmented into training and test subsets according to the global time boundary. This design ensures that samples are temporally ordered and machine-isolated, thereby preventing leakage of future information into past windows.

Importantly, the split also induces a machine-level generalization setting. The S21 machine appears exclusively during the test period and is not present in the training data. Consequently, the model must generalize not only across future market regimes but also to previously unseen hardware configurations. This setup evaluates the robustness of the proposed framework under realistic deployment conditions, where newly released ASIC models may emerge after the training period.

\paragraph{Expanding-window cross-validation:}
\label{subsec:hyperparameters}

\begin{figure*}
        \centering
        \includegraphics[width=\linewidth]{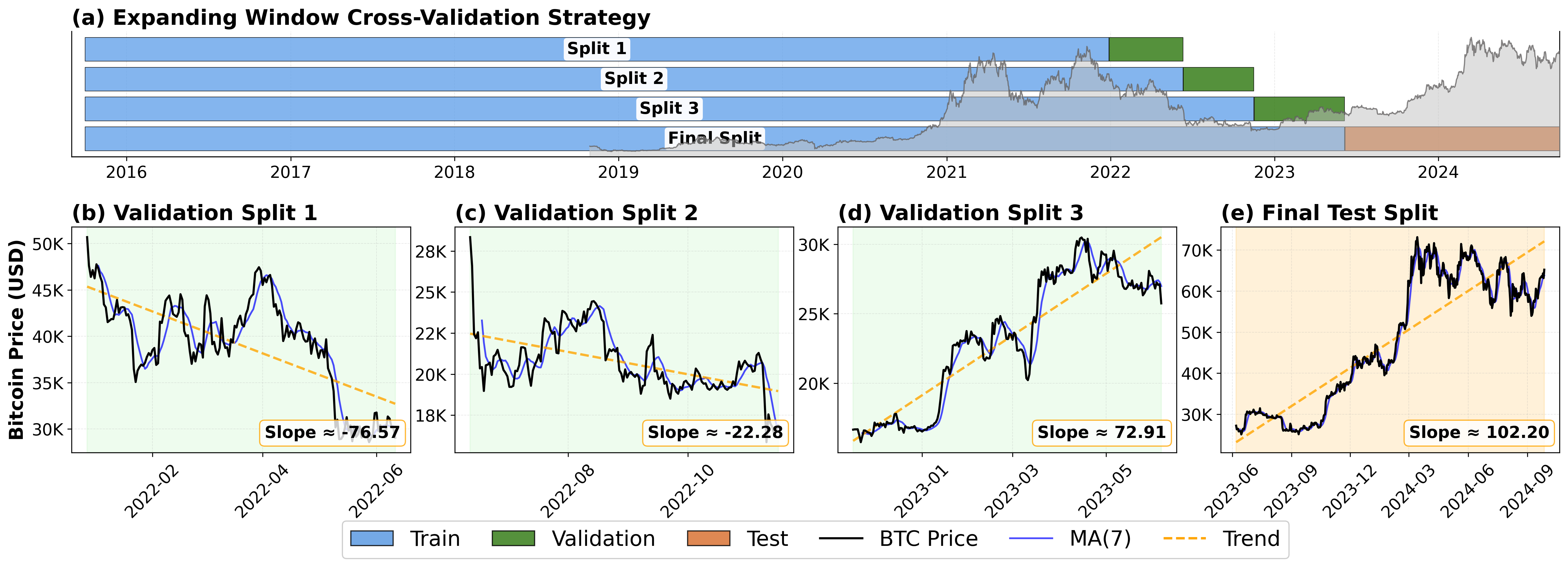}
        \caption{Expanding window cross-validation strategy across market regimes.}
        \label{fig:splits_combined_with_final}
\end{figure*}

Within the training period, we further adopt an expanding-window cross-validation strategy to select hyperparameters and assess robustness under temporally shifting market conditions (Figure~\ref{fig:splits_combined_with_final}(a)). The training timeline is divided into three sequential validation phases (Split~1--3), each preserving chronological order. At each phase $k$, the model is trained on all data up to time $t_k$ and validated on the immediately following segment. This setup mimics a realistic deployment, where only past information is available, and prevents temporal leakage.

Figure~\ref{fig:splits_combined_with_final}(b,c,d) illustrates the progressive expansion of three validation phases: bear market (Split 1), range-bound market (Split 2), and bull market (Split 3). For each model family described in Section~\ref{subsec:baselines}, we perform a grid search over the corresponding hyperparameter space. For each configuration $\theta$, we train and evaluate it on each rolling split and compute macro F1-score on the validation segment to emphasize balanced performance across ROI classes under moderate class imbalance. By evaluating performances across these temporally distinct environments, we ensure that model selection does not overfit to a single regime.

After selecting the best parameters, each model is retrained once on the full training window (October 2015--May 2023) using the same optimization protocol. The final test window (June 2023--September 2024) (Figure~\ref{fig:splits_combined_with_final}(e)) remains strictly unseen during hyperparameter selection and is used only for the final evaluation. The selected configurations for each model are shown in Table~\ref{tab:hyperparameters}.

\begin{table*}[t]
    \centering
    \caption{Selected hyperparameters from rolling expanding-window cross-validation for each model and look-back window. 
Notation: $n_h$-number of heads, $n_l$-number of layers, $lr$-learning rate, $d$-dropout, $d_{feed}$-feedforwarfd dimension, $emb_d$-embedding dimension, $dep$-depth, and $p$-patch size.}
    \label{tab:hyperparameters}
    \begin{tabular}{cll}
        \toprule
        Look back Window & Model & Best model parameters \\
        \midrule
        \multirow{5}{*}{30} 
            & Vanilla LSTM          & $n_h = 8$, $n_l = 4$, $lr = 0.0005$, $d = 0.1$ \\
            & LSTM with blocks      & $n_h = 8$, $n_l = 4$, $lr = 0.001$, $d = 0.2$ \\
            & Vanilla Transformer   & $n_h = 8$, $n_l = 6$, $lr = 0.0001$, $d = 0.2$, $d_{feed} = 256$ \\
            & TSLANet               & $emb_d = 32$, $dep = 6$, $p = 8$, $d = 0.1$, $lr = 0.0005$ \\
            & \textsc{MineROI-Net}  & $n_h = 4$, $n_l = 4$, $lr = 0.001$, $d = 0.3$, $d_{feed} = 256$ \\
        \midrule
        \multirow{5}{*}{60} 
            & Vanilla LSTM          & $n_h = 8$, $n_l = 6$, $lr = 0.001$, $d = 0.1$ \\
            & LSTM with blocks      & $n_h = 16$, $n_l = 4$, $lr = 0.001$, $d = 0.1$ \\
            & Vanilla Transformer   & $n_h = 2$, $n_l = 6$, $lr = 0.0001$, $d = 0.1$, $d_{feed} = 128$ \\
            & TSLANet               & $emb_d = 32$, $dep = 6$, $p = 8$, $d = 0.1$, $lr = 0.0001$ \\
            & \textsc{MineROI-Net}  & $n_h = 2$, $n_l = 2$, $lr = 0.001$, $d = 0.2$, $d_{feed} = 128$ \\
        \bottomrule
    \end{tabular}
\end{table*}

\section{
Results
}
\label{sec:Results}
We report all main comparisons on a strictly held-out test period (June 2023--September 2024), which is not accessed during hyperparameter selection. Hyperparameters for each model are selected using the rolling expanding-window protocol in Section~\ref{subsec:data_split}. After selection, each model is retrained on the full training window (October 2015--June 2023) and evaluated once on the test window. Results are reported as mean $\pm$ standard deviation over five random seeds. We first present the primary setting with a 30-day look-back window, and then analyze the effect of using a longer 60-day history.

\subsection{
Final test performance with 30-day look-back
}
\label{subsec:test_30}
Table~\ref{tab:main_results} summarizes test-set Accuracy and Macro-F1 for all baselines and \model{} model under a 30-day look-back window. This setting serves as our primary configuration because it reflects a practically relevant short-horizon decision context in mining operations, while still capturing near-term dynamics in market price, difficulty, and related covariates.  All values are the mean ± standard deviation over five random seeds.

Across all model families, attention-based approaches outperform purely recurrent sequence modeling. Vanilla Transformer achieves strong performance, indicating that long-range interactions within the 30-day context are informative for ROI classification. Incorporating frequency-aware and cross-feature interaction modules further improves results: LSTM with blocks improves upon Vanilla LSTM, suggesting that explicit spectral feature extraction and channel mixing provide useful inductive bias even with a recurrent backbone.

Overall, \model{} model achieves the best performance on both Accuracy and Macro-F1 in the 30-day setting, outperforming other baseline models. This supports the hypothesis that combining (i) frequency-aware feature extraction, (ii) channel-wise interaction modeling, and (iii) attention-based temporal aggregation yields the most consistent improvements for mining ROI prediction.

\begin{table*}[t]
    \centering
    \caption{Test-set performance (accuracy and macro-F1) with 30-day and 60-day look-back windows (mean $\pm$ std over 5 runs).}
    \label{tab:main_results}
    \begin{tabular}{lcccc}
        \toprule
        & \multicolumn{2}{c}{Look-back window 30} & \multicolumn{2}{c}{Look-back window 60} \\
        \cmidrule(lr){2-3} \cmidrule(lr){4-5}
        Model & Accuracy (Avg$\pm$Std) & F1-score (Avg$\pm$Std) & Accuracy (Avg$\pm$Std) & F1-score (Avg$\pm$Std) \\
        \midrule
        Vanilla LSTM        & 0.611 $\pm$ 0.040 & 0.606 $\pm$ 0.039 & 0.670 $\pm$ 0.036 & 0.669 $\pm$ 0.036 \\
        LSTM with blocks    & 0.637 $\pm$ 0.028 & 0.633 $\pm$ 0.035 & 0.717 $\pm$ 0.081 & 0.718 $\pm$ 0.083 \\
        Vanilla-Transformer & 0.770 $\pm$ 0.045 & 0.769 $\pm$ 0.047 & 0.798 $\pm$ 0.029 & 0.799 $\pm$ 0.029 \\
        TSLANet             & 0.714 $\pm$ 0.058 & 0.716 $\pm$ 0.061 & 0.784 $\pm$ 0.028 & 0.790 $\pm$ 0.028 \\
        \textsc{MineROI-Net} & \textbf{0.785 $\pm$ 0.053} & \textbf{0.790 $\pm$ 0.053} & \textbf{0.832 $\pm$ 0.026} & \textbf{0.835 $\pm$ 0.025} \\
        \bottomrule
    \end{tabular}
\end{table*}

\subsection{
Influence of the look-back window
}
\label{subsec:lookback}
We study the impact of historical context length by varying the look-back window from 30 to 60 days while keeping the evaluation protocol unchanged: hyperparameters are selected using the rolling expanding-window strategy (Section~\ref{subsec:hyperparameters}), after which each model is retrained on the full training period and evaluated once on the strictly held-out test set. Table~\ref{tab:main_results} reports the final test performance for both window lengths.

Overall, increasing the look-back window from 30 to 60 days improves performance across all model families, suggesting that ROI-relevant signals extend beyond a single month. The gains are particularly pronounced for the LSTM-based baselines: Vanilla LSTM increases from 0.611 to 0.670 in accuracy (0.606 to 0.669 in macro-F1), while LSTM with blocks improves from 0.637 to 0.717 in accuracy (0.633 to 0.718 in macro-F1). This indicates that longer histories help recurrent models stabilize their sequential representations, and that the frequency-aware blocks further enhance this benefit.

Attention-based models also benefit from longer context. Vanilla Transformer improves from 0.770 to 0.798 in accuracy (0.769 to 0.799 in macro-F1), reflecting that additional context provides more evidence for the attention mechanism to disambiguate ROI drivers. TSLANet similarly improves from 0.714 to 0.784 in accuracy (0.716 to 0.790 in macro-F1), consistent with the value of longer temporal evidence when combined with convolutional inductive bias and spectral filtering.

Finally, \model{} achieves the best performance across both settings and remains robust as the context length increases. Its accuracy improves from 0.785 to 0.832 and macro-F1 from 0.790 to 0.835, outperforming other baseline models at both window lengths. These results suggest that longer histories provide complementary information to \model{} model's spectral feature extraction and channel-mixing modules, while the Transformer backbone effectively aggregates the extended temporal evidence.

\subsection{
Error analysis and class-wise performance
}
\label{subsec:Error_analysis}
\begin{figure*}
\centering
\begin{subfigure}[b]{0.38\linewidth}
    \centering
    \includegraphics[width=\linewidth]{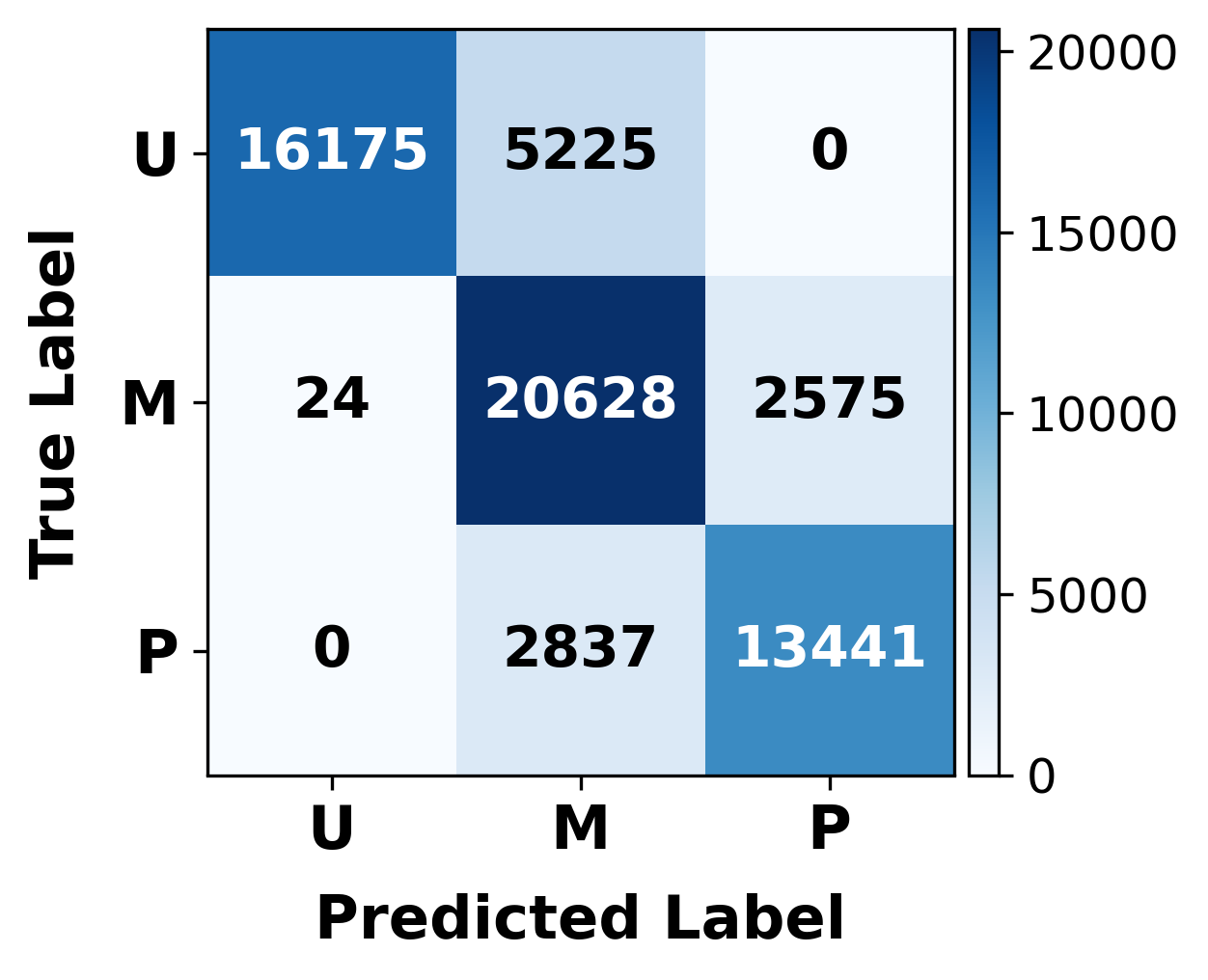}
    \caption{30-day look-back window}
    \label{fig:confusion30}
\end{subfigure}%
\hspace{1.4cm}
\begin{subfigure}[b]{0.38\linewidth}
    \centering
    \includegraphics[width=\linewidth]{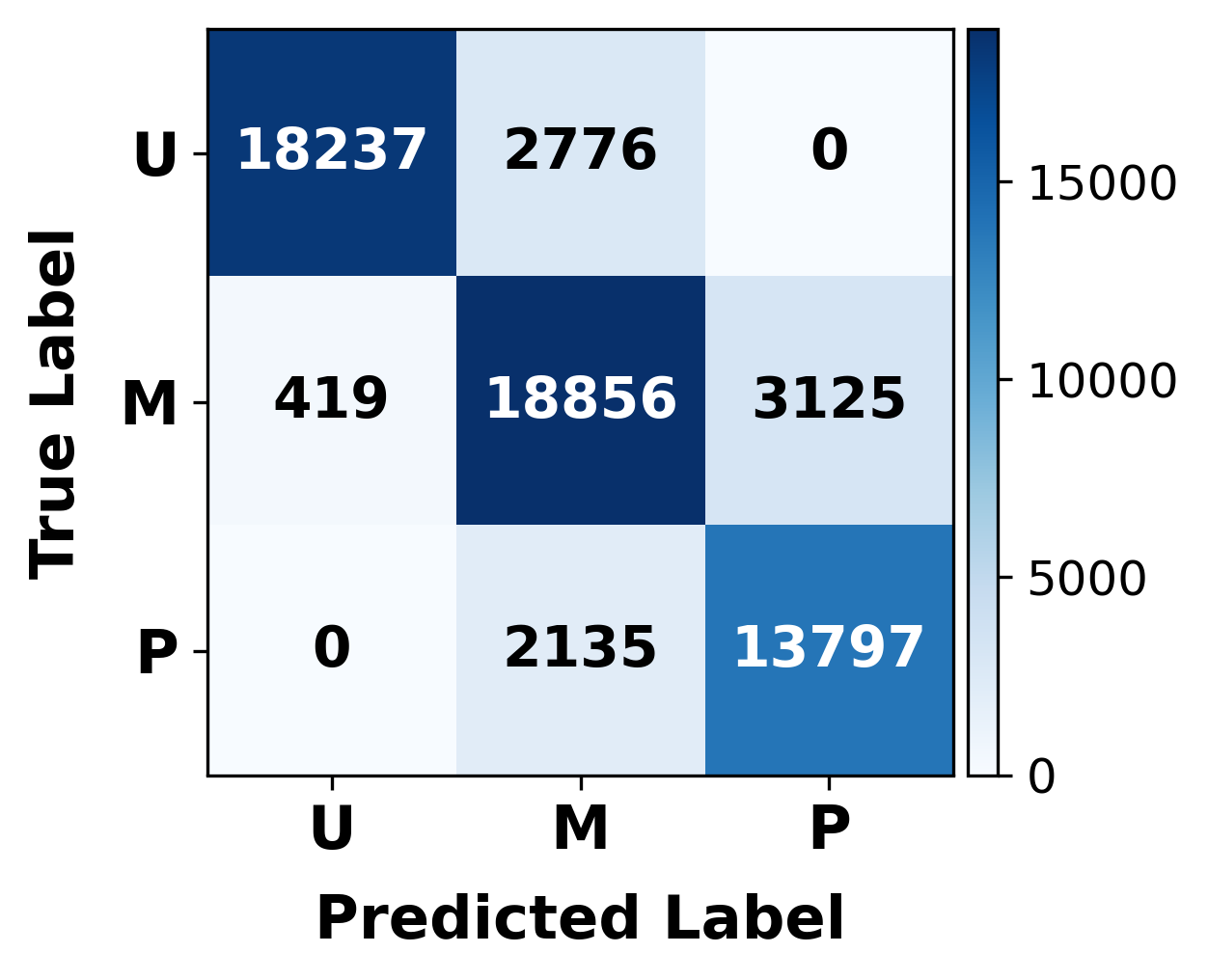}
    \caption{60-day look-back window}
    \label{fig:confusion60}
\end{subfigure}
\caption{Confusion matrices for \model{}.}
\label{fig:confusion}
\end{figure*}

Figures~\ref{fig:confusion30} and~\ref{fig:confusion60} present the confusion matrices of \model{} under 30-day and 60-day look-back windows, and the corresponding per-class metrics are reported in Table~\ref{tab:class_wise}. Across both settings, the model exhibits a clear separation between the extreme classes: importantly, there are no direct confusions between Unprofitable (U) and Profitable (P) (i.e., no U$\rightarrow$P or P$\rightarrow$U predictions). This is desirable for mining hardware decisions, as it reduces the risk of issuing strongly incorrect buy or avoid signals.

\paragraph{Unprofitable (U).} The model attains very high precision for U in both settings (0.999 at 30-day; 0.978 at 60-day), indicating that predicted U cases are highly reliable. The main errors arise from U being predicted as Marginal (M), which is reduced substantially when moving from 30-day to 60-day. Consistently, recall for U improves from 0.756 to 0.868 and the U-class F1-score improves from 0.860 to 0.919 (Table~\ref{tab:class_wise}). Practically, this means the 60-day model flags a larger fraction of truly risky purchases while maintaining very low false alarms.

\paragraph{Profitable (P).} For P, the model achieves strong and more balanced precision--recall trade-offs, improving with longer context. With a 60-day window, recall increases from 0.826 to 0.866 while precision remains high (0.839 $\rightarrow$ 0.815), yielding a higher F1-score (0.832 $\rightarrow$ 0.840). Most residual errors correspond to P being predicted as M (2837 at 30-day vs.\ 2135 at 60-day) (Figures~\ref{fig:confusion30},~\ref{fig:confusion60}), reflecting borderline cases near the profitability boundary rather than catastrophic mistakes.

\paragraph{Marginal (M) as an uncertainty buffer.} The marginal class remains the most challenging, as expected, because it lies near the decision boundary between clear avoid and buy cases. In both settings, the dominant confusions involve M being predicted as P (2575 at 30-day; 3125 at 60-day) and, to a lesser extent, M being predicted as U (24 at 30-day; 419 at 60-day). Despite this, M maintains competitive class-wise performance, with F1-score improving from 0.795 to 0.817 as the look-back window increases. Interpreting M as a conservative ``uncertainty buffer'' is therefore reasonable: ambiguous scenarios can be routed to additional checks (e.g., manual review), while the model’s extreme-class predictions remain reliable.

Overall, the longer 60-day context improves macro-averaged metrics (Macro-F1: 0.829 $\rightarrow$ 0.859; Macro-AUC: 0.881 $\rightarrow$ 0.916), indicating better balanced separability across classes (Table~\ref{tab:class_wise}). Taken together with the absence of U$\leftrightarrow$P confusions, these results suggest that \model{} provides actionable guidance: it reliably identifies high-risk unprofitable scenarios and high-confidence profitable opportunities, while using the marginal class to absorb boundary uncertainty.

\begin{table*}[t]
    \centering
    \caption{Per-class precision, recall, F1-score, and AUC for \model{} with 30-day and 60-day look-back windows on the test set.}
    \label{tab:class_wise}
    \begin{tabular}{lcccccccc}
        \toprule
        & \multicolumn{4}{c}{Look-back 30} & \multicolumn{4}{c}{Look-back 60} \\
        \cmidrule(lr){2-5}\cmidrule(lr){6-9}
        Class & Precision & Recall & F1-score & AUC & Precision & Recall & F1-score & AUC \\
        \midrule
        0 (Unprofitable) & 0.999 & 0.756 & 0.860 & 0.906 & 0.978 & 0.868 & 0.919 & 0.956 \\
        1 (Marginal)     & 0.719 & 0.888 & 0.795 & 0.824 & 0.793 & 0.842 & 0.817 & 0.886 \\
        2 (Profitable)   & 0.839 & 0.826 & 0.832 & 0.913 & 0.815 & 0.866 & 0.840 & 0.907 \\
        \midrule
        Macro Average    & 0.852 & 0.823 & 0.829 & 0.881 & \textbf{0.862} & \textbf{0.859} & \textbf{0.859} & \textbf{0.916} \\
        \bottomrule
    \end{tabular}
\end{table*}

\subsection{
Ablation study of spectral modulation and channel mixing
}
\label{subsec:Ablation}
We conduct an ablation study to quantify the contribution of the two key components in \model{}: (i) spectral modulation and (ii) channel mixing. The benefit of these modules is already visible in the main results (Table~\ref{tab:main_results}): LSTM with blocks, which augments Vanilla LSTM with spectral modulation and channel mixing, improves macro-F1 from 0.669 to 0.718 under the 60-day setting. However, since \model{} achieves the strongest overall performance with a Transformer backbone, we perform a more detailed ablation within this architecture to isolate the individual and combined contributions of each module. Since the 60-day configuration achieves the strongest overall test performance, we perform ablations under this best-performing setting and evaluate all variants on the same held-out test set.

\begin{table*}[t]
    \centering
    \caption{Ablation study of \model{} under the best-performing configuration (60-day look-back) on the held-out test set (June 2023--September 2024). ``Vanilla'' removes both spectral modulation and channel mixing; the other variants add one component at a time. Results are reported as Accuracy and Macro-F1 under seed 42.}
    \begin{tabular}{lcc}\toprule
         \textbf{Variant} &  \textbf{Accuracy}& \textbf{Macro-F1 score} \\\midrule
         Vanilla (w/o Spectral modulation + Channel Mixing) &  0.765& 0.769\\
         + Spectral modulation only                          &  0.785& 0.786\\
         + Channel Mixing only                    &  0.800& 0.804\\
         Full Model (Spectral modulation + Channel Mixing)   &  \textbf{0.858}& \textbf{0.859}\\ \bottomrule
    \end{tabular}
    \label{tab:ablation}
\end{table*}

Table~\ref{tab:ablation} reports test-set Accuracy and macro-F1 for four variants. Removing both modules (Vanilla) yields a macro-F1 of 0.769. Adding spectral modulation alone improves performance to 0.786 macro-F1 (+0.017), indicating that frequency-domain processing provides useful inductive bias for mining ROI classification. Adding the channel mixing module alone yields a larger gain, reaching 0.804 macro-F1 (+0.035), suggesting that explicitly modeling cross-feature interactions is particularly beneficial in this setting, likely due to the heterogeneous drivers of mining profitability.

Combining both components in the full model produces the best results (0.859 macro-F1), substantially outperforming the vanilla backbone (+0.090 macro-F1) and each single-module variant (+0.073 over spectral modulation only and +0.055 over channel mixing only). This indicates that spectral modulation and channel mixing provide complementary benefits: spectral modulation highlights informative periodicities and regime-related frequency content, while channel mixing improves the integration of multi-source signals before temporal aggregation. Overall, the ablation supports our design choice of jointly incorporating frequency-aware modulation and cross-feature interaction modeling in \model{} model.

\subsection{
Sensitivity to alternative conversion-frequency assumptions
}
\label{subsec:selling_strategies}
Our main ROI formulation assumes that mined bitcoin is converted to fiat currency daily at the prevailing BTC/USD price, reflecting a simple operational policy for covering ongoing costs. In practice, however, miners may convert accumulated Bitcoin less frequently, for example, on a weekly or monthly basis, depending on treasury policy and operational constraints. To assess the sensitivity of our results to this modelling choice, we conduct an additional analysis under an alternative monthly conversion rule. Importantly, changing the conversion frequency alters the cash-flow process used to compute ROI and, therefore, the class labels themselves. Rather than retraining a separate model under the monthly rule, we perform a counterfactual relabelling analysis: we keep the model trained under the daily-conversion setting fixed, recompute the ground-truth ROI labels using monthly conversion, and then compare the fixed model’s predictions against these alternative labels. 

Formally, let $(X_{i}, {y_{i}}^{daily})$ denote the original dataset constructed under daily conversion, and let ${y_{i}}^{monthly}$ denote the corresponding labels obtained by recomputing ROI under monthly conversion while keeping the same input sequences $X_i$. We then report performance of the fixed model $f(.)$ as
\begin{equation}
    \hat{y}_i = f(X_i),     \ \text{    and evaluate } \hat{y}_i \text{ against } y_i^{\text{monthly}}. 
    \label{eq:placeholder_label}
\end{equation}

This test quantifies how stable the learned decision boundary is when the downstream economic definition of ROI changes. High agreement indicates that the model captures regime and profitability related structure that remains informative across liquidation policies, and performance degradation indicates that liquidation frequency materially changes class boundaries, motivating the development of training strategy-specific models.

\begin{table}
    \centering
    \caption{Per-class precision, recall, F1-score, and AUC for \model{} with 30-day and 60-day look-back windows on the test set.}
    \begin{tabular}{cccc}\toprule
         &  Precision &  Recall & F1-score \\\midrule
         0 (Unprofitable) &  0.959   &  0.885& 0.921\\
         1 (Marginal)     &  0.786   &  0.839& 0.811\\
         2 (Profitable)   &  0.833   &  0.836& 0.834\\\midrule
         Macro Average &  0.859   &  0.853& 0.856\\ \bottomrule
    \end{tabular}
    
    \label{tab:monthly_table}
\end{table}

Under the original daily-conversion labels, the model achieves an accuracy of 0.858 and a macro-F1 score of 0.859 on the test set (Table ~\ref{tab:class_wise}). When evaluated against the monthly-conversion labels, performance decreases slightly to an accuracy of 0.854 and macro-F1 score of 0.856  (Table ~\ref{tab:monthly_table}), indicating that the learned decision boundary remains largely stable under this alternative settlement assumption.

\begin{table*}[h!]
\centering
\caption{Daily-to-monthly ROI label transitions on the test set. Rows correspond to labels computed under the daily-conversion rule, and columns correspond to labels recomputed under the monthly-conversion rule. Each entry reports the count and row-normalized percentage.}
\label{tab:daily_monthly_transition}
\renewcommand{\arraystretch}{1.2}
\setlength{\tabcolsep}{8pt}
\begin{tabular}{lccc}
\toprule
\textbf{} & \textbf{Monthly U} & \textbf{Monthly M} & \textbf{Monthly P} \\
\midrule
\textbf{Daily U} & 96.3\% & 3.7\%  & 0.0\% \\
\textbf{Daily M} & 0.2\%    & 95.3\% & 4.5\% \\
\textbf{Daily P} & 0.0\%     & 0.4\%   & 99.6\% \\
\bottomrule
\end{tabular}
\end{table*}

Table 6 provides additional insight into the magnitude of the label-definition shift. The vast majority of samples retain the same label when the ROI definition is changed from daily to monthly conversion: 96.3\% of unprofitable cases remain unprofitable, 95.3\% of marginal cases remain marginal, and 99.6\% of profitable cases remain profitable. Importantly, all observed changes occur only between adjacent classes (U$\leftrightarrow$M or M$\leftrightarrow$P); there are no direct transitions between unprofitable and profitable (U$\leftrightarrow$P). This indicates that the effect of monthly conversion is primarily to perturb borderline cases near the decision threshold rather than to induce economically severe reversals.

The class-wise results are consistent with this interpretation. For the unprofitable class, monthly relabeling reduces precision from 0.978 to 0.959 while recall increases from 0.868 to 0.885 (F1-score: 0.919 $\rightarrow$ 0.921), suggesting that some borderline marginal cases move into the unprofitable category under monthly settlement. For the profitable class, precision increases from 0.815 to 0.833 while recall decreases from 0.866 to 0.836, again reflecting mild boundary shifts rather than structural failure. The marginal class remains the most sensitive, with an F1-score decreasing from 0.817 to 0.811, reflecting increased ambiguity near the decision threshold when labels are recomputed under a different settlement rule.

\section{
Conclusion
}
\label{sec:Conclusion}
This paper presents \model{}, an open source data-driven model\footnote{\url{https://github.com/AMAAI-Lab/MineROI-Net}} for predicting the return on investment (ROI) of Bitcoin mining hardware acquisitions, formulated as a three-class time-series classification problem. To our knowledge, this is the first machine learning approach that directly addresses the timing of mining hardware investments, distinguishing between profitable ($\text{ROI} \geq 1$), marginal ($0 < \text{ROI} < 1$), and unprofitable ($\text{ROI} \leq 0$) outcomes over a fixed one-year horizon.

Evaluated on a dataset covering 20 ASIC miners released between 2015 and 2024, \model{} consistently outperforms all four baselines - Vanilla LSTM, LSTM with blocks, Vanilla Transformer, and TSLANet. Under the best-performing 60-day look-back configuration, it achieves 83.2\% accuracy and a macro F1-score of 83.5\%. The model excels at economically critical decisions, achieving 97.8\% precision for detecting unprofitable periods and 81.5\% for profitable ones, importantly, without ever misclassifying a profitable scenario as unprofitable or vice versa. Notably, the test set includes the S21 machine, which is entirely absent from training, demonstrating that \model{} generalizes across unseen hardware configurations. An ablation study confirms that spectral modulation and channel mixing provide complementary benefits, together contributing +0.090 macro F1-score over the vanilla Transformer backbone. Sensitivity analysis further shows that performance remains stable under monthly Bitcoin conversion assumptions, with accuracy and macro F1-score decreasing only marginally to 85.4\% and 85.6\% respectively.

Natural directions for future work include incorporating alternative selling strategies such as weekly or holding-based conversion into the training objective, examining different investment horizons (such as 6- or 18-month ROI), evaluating the framework on additional cryptocurrency mining ecosystems, and exploring regression-based formulations. Overall, \model{} offers clear practical value for capital-intensive mining decisions while maintaining high precision for risk-sensitive operations, providing a reliable tool for the informed timing of mining hardware investments.



\printcredits

\section*{Acknowledgment}
This work was supported by SUTD’s Kickstart Initiative under grant number SKI 2021\_04\_06, and the Ministry of Education (MOE), Singapore, under grant number MOE-T2EP20124-0014. The authors also acknowledge the use of AI-assisted tools for language refinement.

\bibliographystyle{cas-model2-names}

\bibliography{cas-refs}



\end{document}